\documentclass[a4paper,conference]{IEEEtran}

\usepackage[pdftex]{graphicx}
\usepackage{hyperref}
\hypersetup{
    colorlinks=true,
    linkcolor=blue,
    citecolor=black,
    urlcolor=blue,
    pdftitle={Overleaf Example},
    pdfpagemode=FullScreen,
    }
\usepackage{nicefrac}
\usepackage{multirow}
\usepackage{algpseudocode,algorithm,algorithmicx}
\newcommand*\Let[2]{\State #1 $\gets$ #2}
\algrenewcommand\algorithmicrequire{\textbf{Arguments:}}
\usepackage{xspace}

\newcommand{\dW}{$\frac{dL}{dW}$\xspace}
\newcommand{\dO}{$\frac{dL}{dO}$\xspace}
\newcommand{\dOp}{$\bar{\frac{dL}{dO}}$\xspace}
\newcommand{\dD}{$\frac{dL}{dD}$\xspace}

\begin{document}

\title{Accelerating DNN Training\\with Structured Data Gradient Pruning}

\author{\IEEEauthorblockN{Bradley McDanel}
\IEEEauthorblockA{Franklin \& Marshall College}
\and
\IEEEauthorblockN{Helia Dinh}
\IEEEauthorblockA{Franklin \& Marshall College}
\and
\IEEEauthorblockN{John Magallanes}
\IEEEauthorblockA{Franklin \& Marshall College}}

\maketitle

\begin{abstract}
Weight pruning is a technique to make Deep Neural Network (DNN) inference more computationally efficient by reducing the number of model parameters over the course of training. However, most weight pruning techniques generally does not speed up DNN training and can even require more iterations to reach model convergence. In this work, we propose a novel Structured Data Gradient Pruning (SDGP) method that can speed up training without impacting model convergence. This approach enforces a specific sparsity structure, where only N out of every M elements in a matrix can be nonzero, making it amenable to hardware acceleration. Modern accelerators such as the Nvidia A100 GPU support this type of structured sparsity for 2 nonzeros per 4 elements in a reduction. Assuming hardware support for 2:4 sparsity, our approach can achieve a 15-25\% reduction in total training time without significant impact to performance. Source code and pre-trained models are available at \url{https://github.com/BradMcDanel/sdgp}.
\end{abstract}

\IEEEpeerreviewmaketitle

\section{Introduction}
Deep Neural Networks (DNNs) are now widely used for many applications, such as computer vision, speech recognition, and natural language processing. However, their large number of parameters and associated computational complexity makes DNN training expensive. Most prior research on reducing DNN computation costs focus on DNN inference. For instance, quantization~\cite{lin2016fixed} can reduce the number of bits required to represent model parameters, allowing for more efficient storage and faster execution. Weight pruning~\cite{han2015deep} is another technique that can reduce the number of model parameters, by setting some parameters to zero during training. However, while these approaches can lead to more efficient inference, they often make DNN training less efficient. For instance, weight pruning often requires a longer training regime to reach convergence after multiple rounds of pruning~\cite{frankle2018lottery}.

\begin{figure}
    \centering
    \includegraphics[width=\columnwidth]{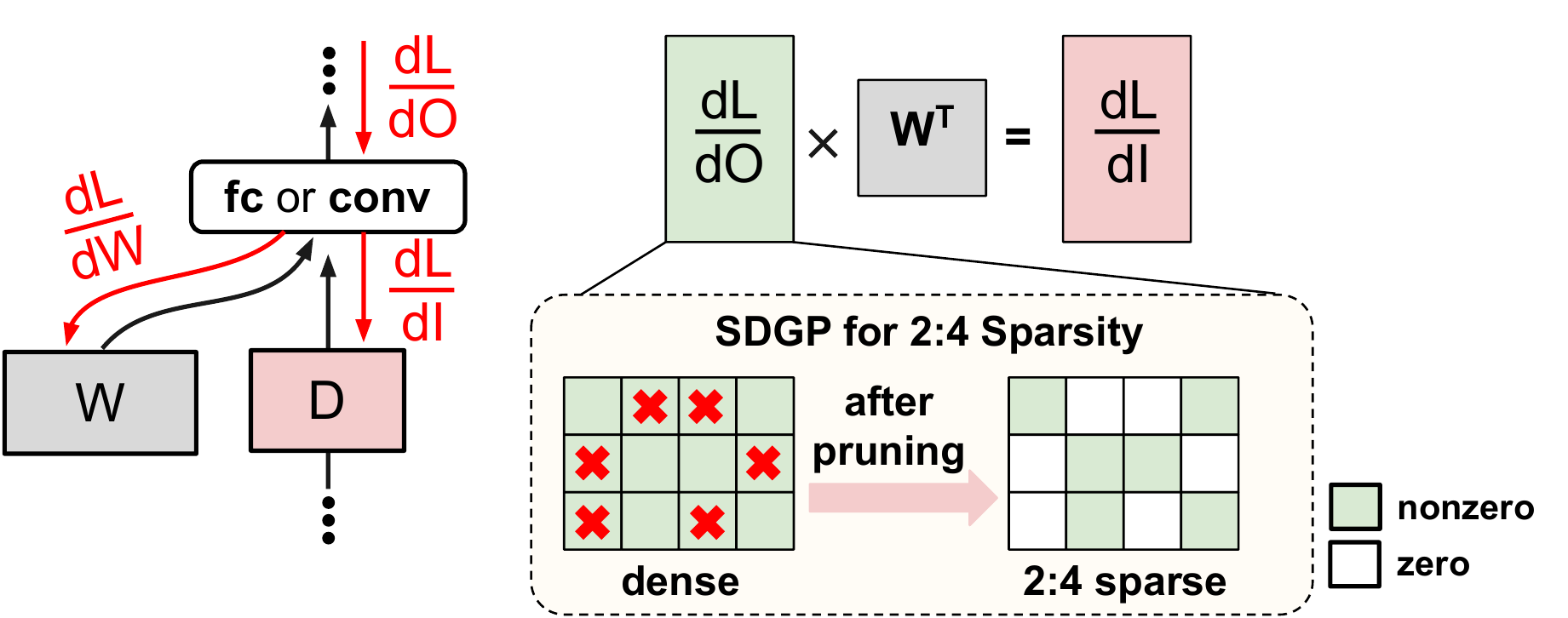}
    \caption{Structured data gradient pruning (SDGP) prunes the output data gradients from the layer above during backpropagation before computing the input data gradients for the layer below. Each group in the dense matrix is pruned such that only $n=2$ nonzeros remain out of $m=4$ total elements. SDGP reduces the number of matrix operations required to compute the input gradients, leading to faster training on support hardware.}
    \label{fig:overview}
\end{figure}

In this work, we propose a novel structured data gradient pruning (SDGP) technique to speed up DNN training. Our approach is based on the observation that modern hardware accelerators, such as the Nvidia A100 GPU, support a specific type of structured sparsity, where only 2 out of every 4 elements in a matrix can be nonzero. We enforce this sparsity structure on the activations gradients during training, in order to achieve a 2$\times$ reduction in operations required to compute the input gradients for the previous layer. Figure~\ref{fig:overview} provides and overview of how SDGP prunes data gradients. During backpropagation, the output data gradients from the preceding layer are pruned by setting some elements to zero, before computing the input data gradients for the layer below. This approach reduces the number of matrix operations required to compute the input gradients, which can speed up training. We find that using SDGP achieves comparable performance (i.e., classification accuracy) to using dense data gradients.

In order for SDGP to minimally impact training stability, the pruned gradients must still closely approximate the original gradients. To this end, we evaluate multiple variants of SDGP that use different ranking mechanism to determine which elements in a group should be pruned. Borrowing from prior work on structured weight pruning~\cite{wen2016learning}, we propose three ranking schemes: (1) random pruning, (2) a greedy ranking scheme that prunes the smallest magnitude elements, and (3) a version of the greedy scheme that adjusts the remaining nonzero elements to better approximate the original distribution.

The A100 only supports a 50\% sparsity ratio (with $n=2$ and $m=4$). In this work, we also evaluate higher sparsity ratios than supported by the A100 (e.g., 87.5\% sparse with $n=2$ and $m=16$) in order to see how the proposed approach scales with increased sparsity. Generally, higher sparsity ratios lead to worse performance, but further reduce the training time. In section~\ref{sec:eval:reduction}, we provide a Time-To-Accuracy~\cite{coleman2017dawnbench} (TTA) on the relative performance a model trained with a given sparsity ratio can achieve in a fixed time budget. 

We evaluate SDGP on a number of deep neural networks, including a 9-layer ResNet~\cite{he2016deep} (denoted ResNet-9) on CIFAR-10~\cite{krizhevsky2014cifar} and ResNet-18, ResNet-50, and RegNet~\cite{radosavovic2020designing} on ImageNet~\cite{deng2009imagenet}. Across all networks, we find that SDGP can speed up training by up to 15-25\% without any significant loss in accuracy. Additionally, since SDGP only prunes data gradients, it can be easily integrated into existing deep learning frameworks without impacting how users construct or train model (similar to mixed-precision training~\cite{micikevicius2018mixed}).

The novel contributions of the paper are:
\begin{itemize}
    \item Structured data gradient pruning (SDGP) which speeds up training without impacting model convergence.
    \item A detailed evaluation of how different ranking schemes used by SDGP affect the pruning process and the final accuracy of the models.
    \item Efficient CUDA kernels for implementing SDGP that introduce minimal overhead during backpropagation.
\end{itemize}

The rest of the paper in organized as follows: Section~\ref{sec:bg} provides background on DNN training and discusses related work on structured pruning, Section~\ref{sec:sdgp} presents the SDGP algorithm and outlines how pruning is performed on data gradients, and Section~\ref{sec:eval} compares SDGP to conventional DNN training and analyzes the impact of the sparsity ratio on model performance and runtime. Finally, we discuss future work and summarize the results in Section~\ref{sec:conc}.

\section{Background and Related Work}
\label{sec:bg}

In this section, we first review the computation of DNN training in Section~\ref{sec:bg:dnn-training}. Then, in Section~\ref{sec:bg:sparse-hw}, we review hardware accelerators that support sparse computations. Finally, we discuss related work on structured weight pruning in Section~\ref{sec:bg:structured-pruning}. 

\subsection{Computation of DNN Training}
\label{sec:bg:dnn-training}

Each DNN training iteration consists of two stages: a forward pass and a backward pass. In the forward pass, the input data is propagated forward through all layers of the network, and a loss function is used to compute the error of the network with respect to some target. In backward pass, the computed error is used to update weights by propagating it backwards through the network. 

\begin{figure}
    \centering
    \includegraphics[width=\columnwidth]{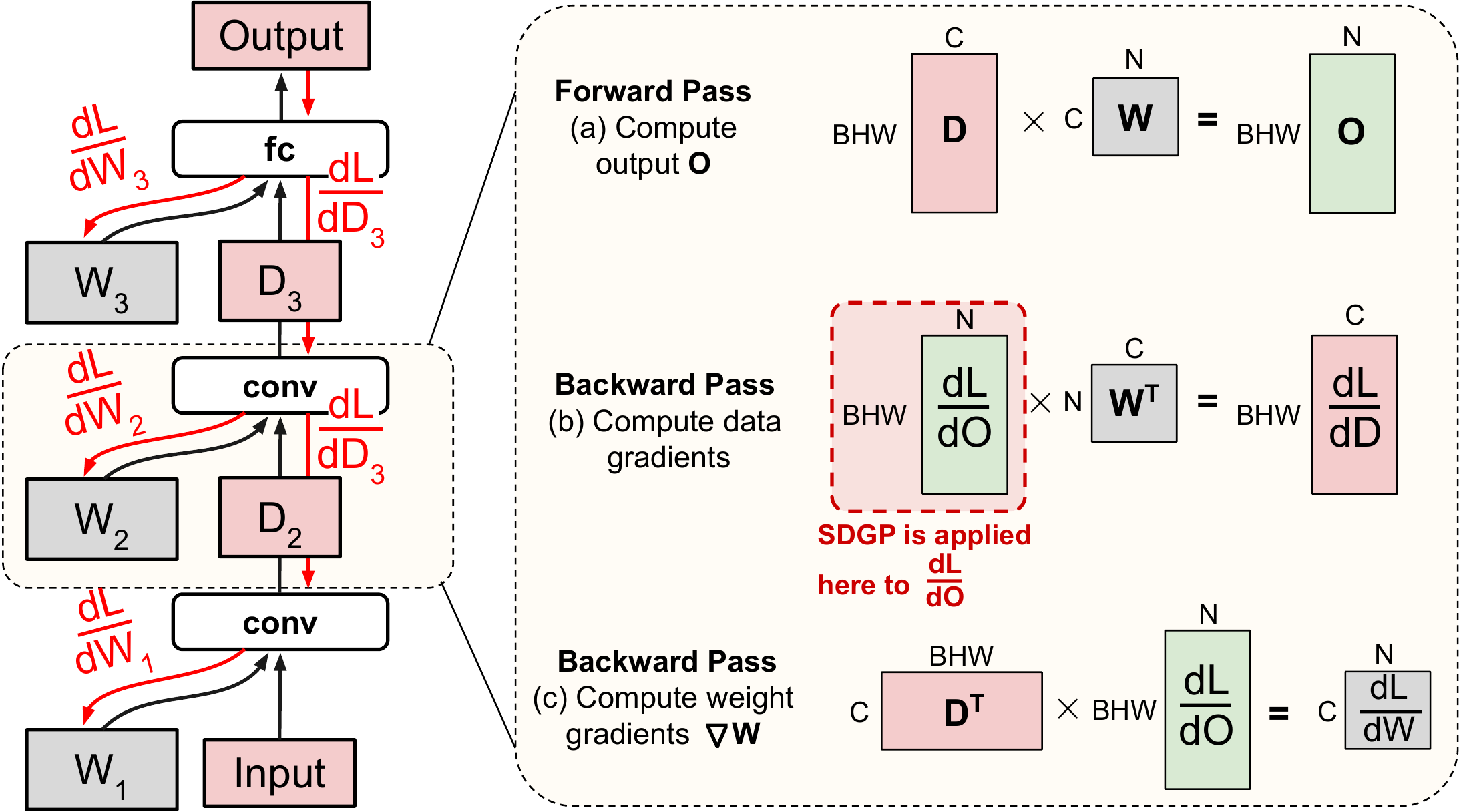}
    \caption{(left) A 3-layer DNN with the forward pass shown using black arrows and the backward pass shown using red arrows. (right) The computation performed for a single layer during the forward and backward pass as matrix multiplications. The proposed SDGP is applied to \dO before computing the data gradient \dD. Normalization and activation layers are omitted for clarity.}
    \label{fig:dnn-training}
\end{figure}

Figure~\ref{fig:dnn-training} provides a computational view of a single training iteration. On the left, we see a 3-layer DNN, with the forward pass shown using black arrows and the backward pass shown using red arrows. During the forward pass, each fully connected or convolutional layer computes a data (activation) tensor that becomes the input for the next layer. During the backward pass, a weight gradient (\dW) and data gradient (\dD) is computed for each layer. The right side of Figure~\ref{fig:dnn-training} illustrates the computations of a single layer during these forward and backward passes as matrix multiplications. For simplicity, we assume the convolutional kernel size is $1\times1$. In general, im2col~\cite{ChetlurWVCTCS14} can be used to map convolution into matrix multiplication. In this work, we propose to prune the output data gradients \dO before computing the input data gradients \dD using SDGP. This pruning operation is performed on every layer in the network in each training iteration.

\subsection{Sparse Hardware Accelerators}
\label{sec:bg:sparse-hw}
The majority of prior work on reducing DNN training and inference runtime has a focus on leveraging sparsity present in weights and activations during forward propagation~\cite{mahmoud2020tensordash,zhang2019eager,yang2020procrustes,choi2020energy}. TensorDash~\cite{mahmoud2020tensordash} automatically skips multiplications with zero values in the activation data, which commonly occurs when using certain activation functions such as ReLU~\cite{nair2010rectified}. Eager Pruning~\cite{zhang2019eager} and Procrustes~\cite{yang2020procrustes} co-designing DNN training to fit a given hardware platform. Column combining~\cite{kung2019packing} adds a structured sparsity constraint that allows only $n$ nonzeros per $m$ elements in a DNN weight matrix and designed a corresponding systolic~\cite{kung1982systolic} to efficiently support sparse structured matrix multiplication. The Nvidia A100 sparse tensor cores introduce a similar N:M sparsity constraint with a different underlying hardware architecture~\cite{Nvidia}.

\subsection{Structured Pruning Techniques}
\label{sec:bg:structured-pruning}
DNN pruning is an extensively studied technique for reducing DNN model size and runtime~\cite{han2015deep,molchanov2016pruning,sanh2020movement,he2017channel,molchanov2016pruning,li2016pruning,frankle2018lottery, liu2018rethinking, zhuang2018discrimination,he2019filter}. As stated earlier, the majority of pruning techniques are applied to weight tensors over the course of training to make models smaller and more efficient when deployed at runtime for inference.

Multiple pruning functions have been proposed to determine the importance of the DNN weights, including magnitude-based pruning~\cite{han2015deep} and gradient-based approaches~\cite{lecun1990optimal, molchanov2016pruning,sanh2020movement}. Much of the initial work on weight pruning did not add any structured constraints, such n:m sparsity, when determining which elements to prune. Generally, such unstructured pruning techniques lead to irregular distributions on nonzero weights, making efficient hardware implementation difficult. By comparison, structured pruning techniques~\cite{wen2016learning,zhuang2018discrimination,ye2018rethinking,he2019filter,luo2020autopruner} add additional constraints that lead to sparse tensors that can be more efficiently processed. Typically, more rigid sparsity structures, such as pruning entire filters, leads to sub-optimal performance (e.g., classification) compared to unstructured pruning but lead to significantly faster inference. The N:M structured sparsity has achieved significantly less attention that these prior methods. Zhou et al. proposed to enforce weight tensors with the N:M sparsity pattern at the start of training~\cite{zhou2021learning}. In~\cite{hubara2021accelerated}, the authors apply a transposable structured pruning mask to the weights to support efficient computation during both forward and backward passes.

In this work, we forgo adding any sparsity constraint to weight tensors and instead enforce N:M structured sparsity on the gradients. Therefore, this approach could potentially be used with any of the proposed weight pruning approaches described above to achieve further savings.

Ye et al. also explore pruning activation gradients~\cite{ye2020accelerating} to reduce computation during backpropagation. However, they use an unstructured pruning algorithm making it difficult to realize the theoretical benefits of training during pruning. 

\section{Structured Data Gradient Pruning}
\label{sec:sdgp}
In this section, we provide an overview of how structured data gradient pruning (SDGP) is used during DNN training. First, in Section~\ref{sec:sdgp:alg}, we show how SDGP is applied to data gradient tensors. Next, Section~\ref{sec:sdgp:mechs}, gives an overview of the different structured pruning we use in our evaluation. Finally, in Section~\ref{sec:sdgp:motiv}, we provide intuition for why SDGP can be applied without significantly impacting model performance.

\subsection{SDGP Algorithm}
\label{sec:sdgp:alg}

Algorithm~\ref{alg:sdgp} shows how structured data gradient pruning is applied to a gradient data tensor $D$. The $n$ and $m$ parameters determine the number of nonzeros per group and group size, respectively. The pruning function $\mathcal{P}$ (discussed next in Section~\ref{sec:sdgp:mechs}) dictates which $n$ element across the group $m$ are kept and prunes the other $m - n$ elements. The algorithm works by simply iterating across all groups, applying $\mathcal{P}$ to each group, and saving the pruned version of the group at the corresponding location in the output data gradient tensor $\hat{O}$. 

\begin{algorithm}
  \caption{Structured Data Gradient Pruning (SDGP)\label{alg:sdgp}}
  \begin{algorithmic}[1]
    \Require{\dO output data gradient tensor, $n$ nonzeros, $m$ group size, $\mathcal{P}$~pruning~function (described in Section~\ref{sec:sdgp:mechs})}
    \Statex
    \Function{sdgp}{\dO, $n$, $m$, $\mathcal{P}$}
      \Let{\dOp}{\texttt{zeros\_like}$(\frac{dL}{dO})$}
      \Comment{\dOp: holds pruned \dO}
      \Let{$k$}{$|\frac{dL}{dO}|$}

      \For{$i \gets 0 \textrm{ to } k$}
        \Let{$\bar{\frac{dL}{dO}}\lbrack im:im+m \rbrack$}{$\mathcal{P}(\frac{dL}{dO}\lbrack im:im+m \rbrack, n)$}
      \EndFor
      \State \Return{\dOp}
    \EndFunction
  \end{algorithmic}
\end{algorithm}

Note that Algorithm~\ref{alg:sdgp} assumes that the pruning dimension is the last dimension in the tensor (meaning that the stride between elements in the group is 1). This simplified the implementation and leads to a more efficient CUDA implementation (see \url{https://github.com/BradMcDanel/sdgp/blob/main/kernels/prune_kernel.cu}). For convolutional layers, we use $N$ as the pruning dimension (as depicted in Figure~\ref{fig:dnn-training}b) which is commonly referred to as the channel dimension. Therefore, in our implementation, we require the channel dimension to be the final dimension. This is generally not an issue as using channels as the final dimension is a commonly used memory layout when using Tensor Cores in Nvidia A100 GPUs~\cite{Nvidia}.

\begin{figure}
    \centering
    \includegraphics[width=\columnwidth]{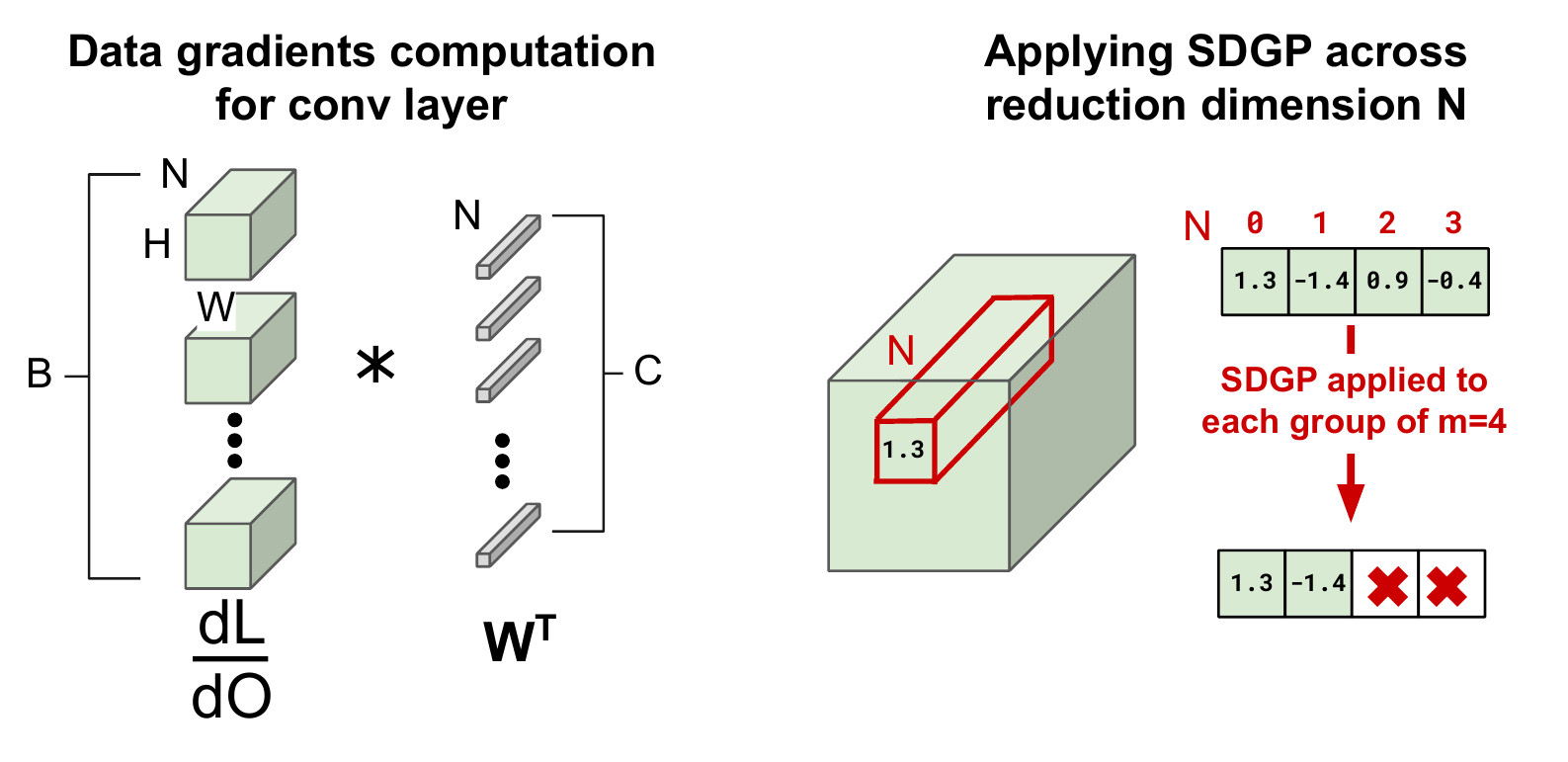}
    \caption{(left) The computation required to compute the data gradients \dD for a convolutional layer. (right) How SDGP is applied to the output data gradients \dO before performing the data gradient computation. Only a single group at a fixed $B$, $W$, and $H$, position is shown.}
    \label{fig:apply-sdgp}
\end{figure}

Figure~\ref{fig:apply-sdgp} shows how SDGP is applied across this pruning dimension of the output data gradients \dO before computing the input data gradients \dD. In our CUDA implementation, each thread is responsible for determine which elements to prune in a given group. The implementation supports a maximum group size of $m=32$. Since \dO is also used to compute the weight gradients \dW, SDGP creates a copy of \dO to store the pruned data gradient tensor \dOp. 

\subsection{Structured Pruning Functions}
\label{sec:sdgp:mechs}
In this section, we present multiple pruning functions used in conjunction with SDGP. Algorithm~\ref{alg:sdgp} requires a pruning function $\mathcal{P}$ to perform the pruning for each group. In this work, we evaluate three pruning function, which can be summarized as follows:
\begin{itemize}
    \item \textbf{Random}: Randomly prunes $m - n$ elements per group.
    \item \textbf{Magnitude}: Sorts the elements in a group by magnitude and prunes the smallest $m - n$ elements per group.
    \item \textbf{Rescaled Magnitude}: Uses magnitude pruning, then rescales the remaining $n$ elements such that\\$||D||_1 = ||\hat{D}||_1$.
\end{itemize}

\begin{figure}
    \centering
    \includegraphics[width=\columnwidth]{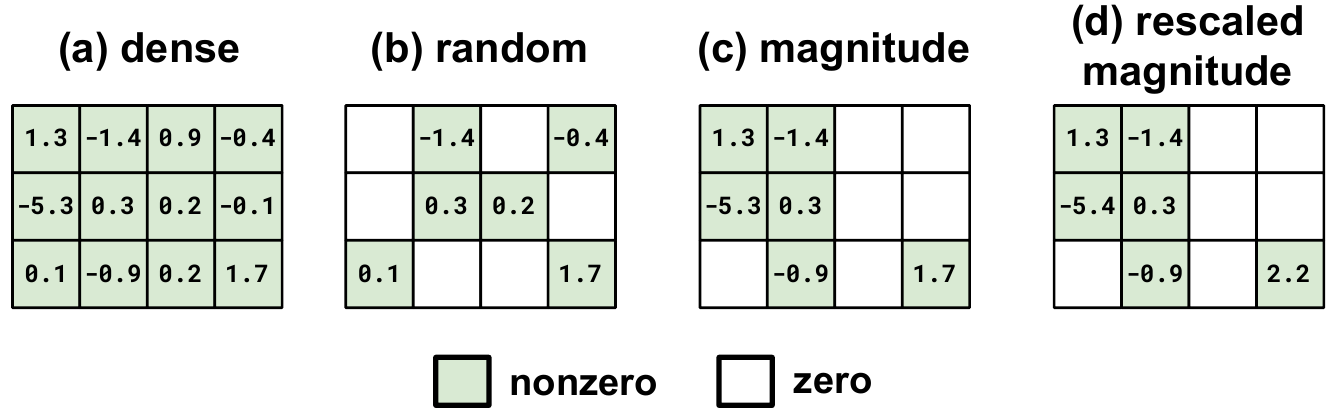}
    \caption{(a) Dense gradient tensor. (b)-(d) Apply random pruning, magnitude pruning, and rescaled magnitude pruning to (a). Note that (d) rescales the nonzero values after pruning such that the $L^1$-norm is preserved across batch and spatial dimensions per channel.}
    \label{fig:prune-algs}
\end{figure}

Figure~\ref{fig:prune-algs} shows the result of each pruning function applied to an input dense gradient tensor. For simplicity, only 3 rows, representing batch and spatial dimensions (i.e., width and height for images), and 4 columns, representing channels, are depicted. In practice, one of the these pruning algorithms will be applied to each $m$ element group across all network layers. As discussed later in Section~\ref{sec:eval:comp}, we find that rescaling the remaining nonzeros stabilize training and leads to slightly higher accuracy compared to simple magnitude pruning.

To support magnitude pruning at the group-level, we must first sort all $m$ elements in the group by their magnitude and then select the top $n$ to keep. For instance, for the group $\lbrack 1, 7, -3, 2 \rbrack$, we first rank the elements by magnitude: $\lbrack 7, -3, 2, 1 \rbrack$. Then, we prune the smallest $m - n$. So, for $n=2$, the group would become $\lbrack 7, -3, 0, 0 \rbrack$. Importantly, we must store the original positions of each element in the group before ranking in order to restore the remaining elements to their original position: $\lbrack 0, 7, -3, 0 \rbrack$. All of these steps are performed by a single CUDA thread that is responsible for a given group.

\subsection{Motivating SDGP via Data Gradient Distribution}
\label{sec:sdgp:motiv}
A conventional argument from weight pruning methodologies is that smaller weights contribute less to the resulting activation values, making them less useful than larger weights from a computational standpoint~\cite{han2015deep}. In this section, we similarly argue that extremely small gradients will have minimal impact on the resulting weight gradients and therefore convergence of the model.

\begin{figure}
    \includegraphics[width=\columnwidth]{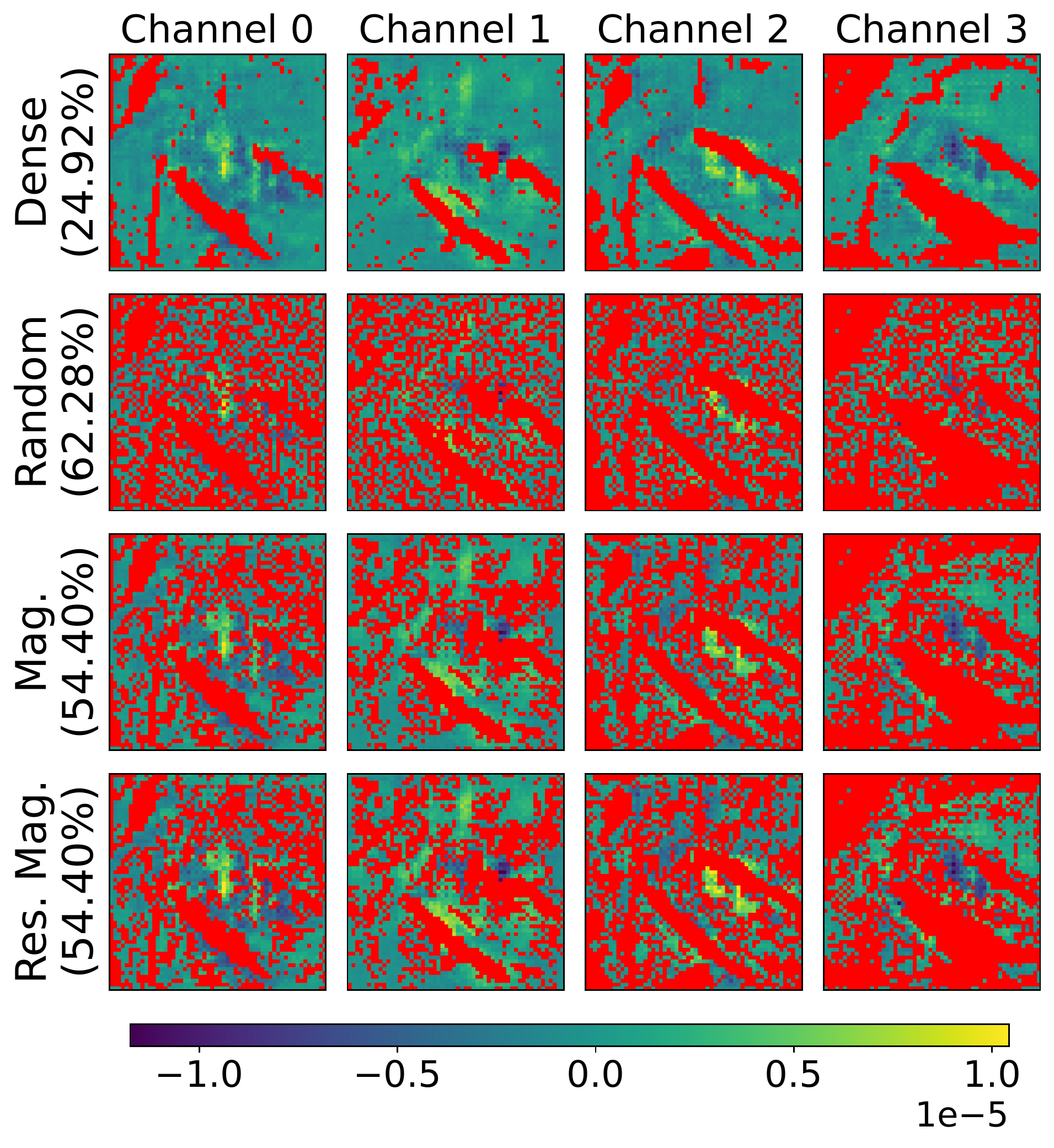}
    \caption{(top) Data gradients across four neighboring (i.e., grouped) channels. The red pixels denote exact 0 values. The remaining rows show the impact of the various pruning functions on the group of channels.}
    \label{fig:gradient-dist}
\end{figure}

Figure~\ref{fig:gradient-dist} (top row) shows four neighboring data gradient channels for the second convolutional layer of ResNet-18 trained on ImageNet after the first epoch of training. The red pixels denote exact zero values that have potential to be skipped under the structured pruning constraint. The reason that almost 25\% of values in these data gradients are zero is due to the use of 16-bit floating point representations, which is supported natively by the A100 and roughly halves the training time compared to more conventional 32-bit floating point.

The remaining three rows of Figure~\ref{fig:gradient-dist} show the impact of applying the different pruning algorithms proposed in Section~\ref{sec:sdgp:mechs} on the top row. For all settings, two nonzeros $n=2$ are kept across each group of $m=4$ elements. As the pruning is performed across the channel dimension, only two of the four elements at each spatial position can be nonzero.

Random pruning often prunes high magnitude regions, which lead to lower performance as we discuss in Section~\ref{sec:eval}. Magnitude and rescaled magnitude pruning always prune the same low-magnitude values. However, conventional magnitude pruning biases the distribution towards zero as it only sets values to zero. By comparison, the rescaling operation used by the rescaled magnitude pruning function preserves the $L^1$-norm across batch and spatial dimensions per channel.

\section{Evaluation}
\label{sec:eval}
In this section, we evaluate the performance of SDGP on a variety of DNNs and datasets to illustrate its robustness across domains. First, in Section~\ref{sec:eval:comp}, we compare the performance of conventional DNN training (using dense activation gradients) to SDGP with the various pruning methods outlined in Section~\ref{sec:sdgp:mechs}. Then, in Section~\ref{sec:eval:sparsity}, we analyze the impact of the degree of sparsity in SDGP (e.g., $m=4$ versus $m=8$ for a fixed $n=2$) on accuracy. Finally, in Section~\ref{sec:eval:reduction}, we estimate the reduction in training time for a variety of SDGP sparsity settings.

We evaluate SDGP on CIFAR-10 using a 9-layer version of ResNet~\cite{he2016deep} and ImageNet~\cite{deng2009imagenet} using ResNet-18, ResNet-50, and RegNetX-400MF~\cite{radosavovic2020designing}. We use the FFCV~\cite{leclerc2022ffcv} dataloader and training regime to train all networks. For CIFAR-10 and ImageNet, all networks are training for 150 and 88 epochs, respectively. Following FFCV, we  employ a cyclic learning rate~\cite{smith2017cyclical}, a resolution scaling regime, label smoothing~\cite{szegedy2016rethinking}, weight decay, and several other regularization techniques. These techniques are applied equally across all SDGP settings to provide a fair comparison. All models were trained using either 1 or 4 Nvidia A100 GPUs. A complete set of training settings are provided in configuration scripts in the code repository. SDGP is applied to all convolution layers in each training iteration.

\subsection{Comparison with Conventional DNN Training}
\label{sec:eval:comp}
\begin{figure}
    \includegraphics[width=\columnwidth]{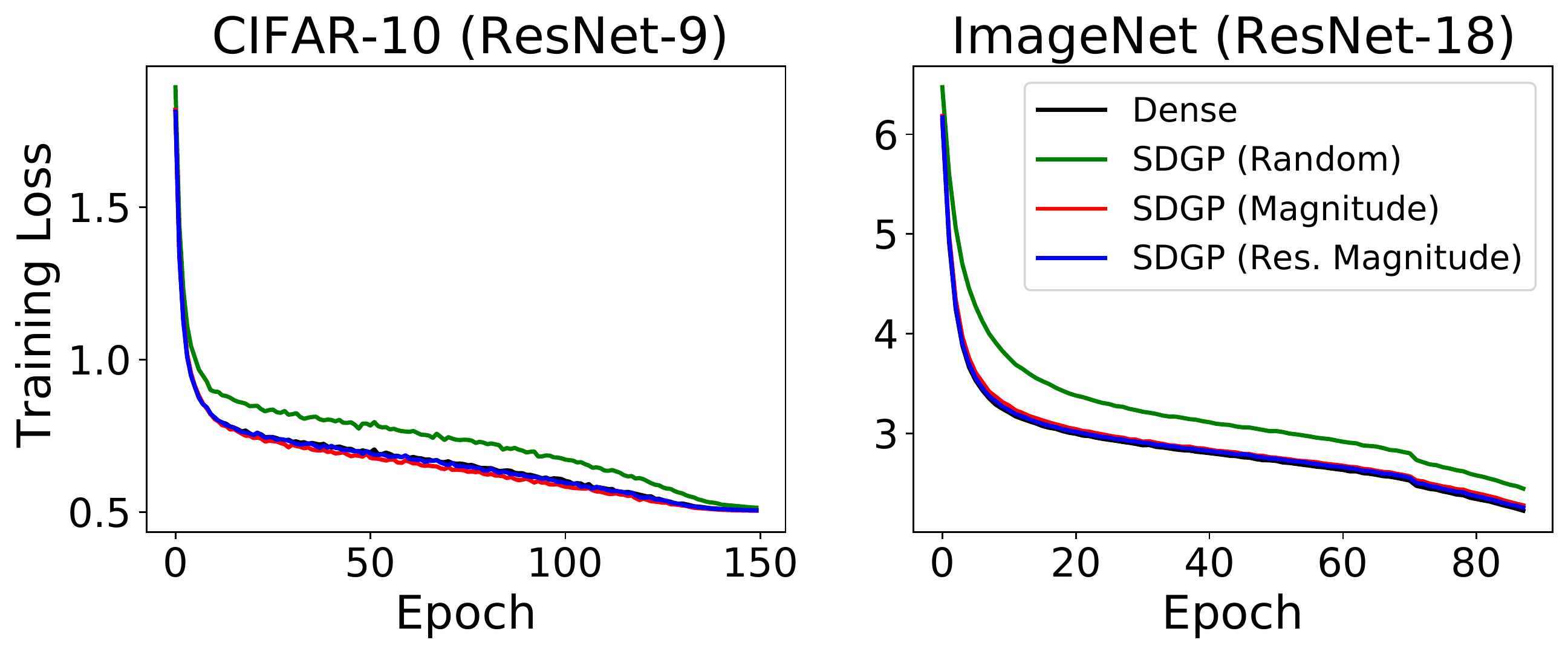}
    \caption{Training loss for CIFAR-10 and Imagenet.}
    \label{fig:training-loss}
\end{figure}

In this section, we compare the classification accuracy of conventional DNN training against SDGP using the different pruning techniques outlines in Section~\ref{sec:sdgp:mechs}. Figure~\ref{fig:training-loss} show the average softmax cross entropy loss for the training dataset over all training epochs for CIFAR-10 and ImageNet. For all SDGP settings, number of nonzeros per group $n=2$ and group size $m=4$. This configuration matches the structured sparsity supported by the sparse tensor cores in the Nvidia A100~\cite{Nvidia}. The SDGP (Random) loss curves are generally higher then the dense model. By comparison, SDGP (Magnitude) and SDGP (Res. Magnitude) have similar loss curves to the baseline (Dense) setting.

\begin{figure}
    \includegraphics[width=\columnwidth]{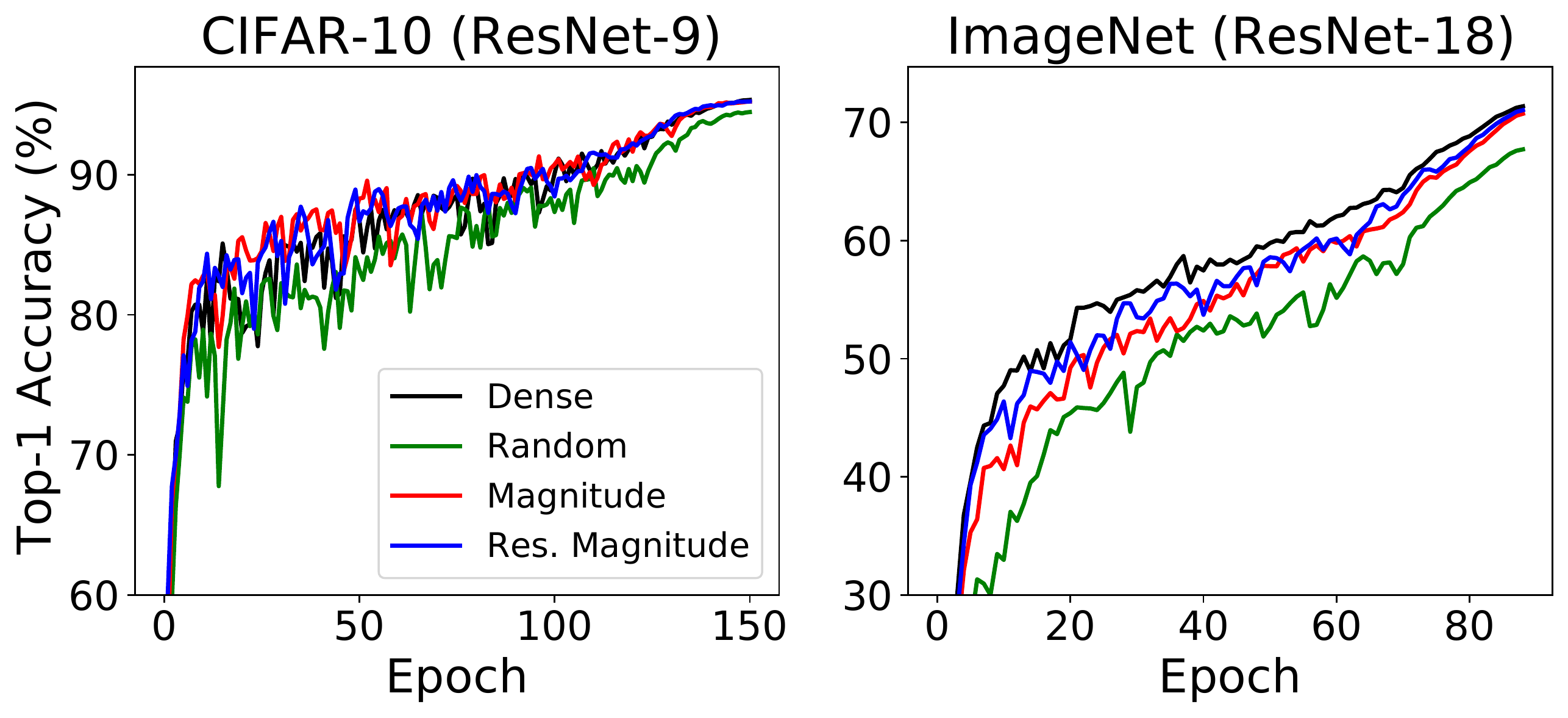}
    \caption{Top-1 validation accuracy (\%) for CIFAR-10 and ImageNet.}
    \label{fig:training-acc}
\end{figure}

Figure~\ref{fig:training-acc} shows the top-1 validation accuracy for the same models depicted in Figure~\ref{fig:training-loss}. The SDGP (Random) settings achieve the lowest classification accuracy. This is likely due to the fact that randomly pruning gradients leads to many larger important gradients from being pruned more often, which can be seen in Figure~\ref{fig:gradient-dist} (second row). By comparison, both magnitude-based methods, SDGP (Magnitude) and SDGP (Rescaled Magnitude), maintain a similar classification accuracy as the baseline setting throughout the course of training. Note that the $n=2$ and $m=4$ structure sparsity supported by the A100 is a relatively conservative setting in terms of achieved sparsity (50\%). In Section~\ref{sec:eval:sparsity}, we explore the trade-off between sparsity and classification accuracy.

Table~\ref{tab:accuracy} shows the final top-1 validation accuracy for all evaluated DNNs. Generally, we see that SDGP is able to achieve similar accuracy as the baseline setting across all DNNs. This suggests that users may be able to use SDGP as a drop-in replacement without noticing a significant impact in performance. The benefit of pruning data gradients, as opposed to pruning weights/data, is that the resulting model does not require any additional considerations both during training and when deploying the model for inference. Even though the relative reduction in training time is modest (15-25\%), it makes no modifications to the model and minimal changes to the training framework.

\begin{table}
\caption{Top-1 Validation Accuracy for Baseline (Dense)\\and SDGP (Rnd, Mag, and Res Mag).}
\begin{center}
\begin{tabular}{cc|cccc|}
\cline{3-6}
\multicolumn{1}{l}{}                            & \multicolumn{1}{l|}{} & \multicolumn{4}{c|}{Top-1 Validation Accuracy (\%)}                                                    \\ \hline
\multicolumn{1}{|c|}{Dataset}                   & DNN Model             & \multicolumn{1}{c|}{Dense} & \multicolumn{1}{c|}{Rnd.} & \multicolumn{1}{c|}{Mag.} & Res. Mag. \\ \hline
\multicolumn{1}{|c|}{CIFAR-10}                  & ResNet-9              & \multicolumn{1}{c|}{95.3}  & \multicolumn{1}{c|}{94.5} & \multicolumn{1}{c|}{95.2} & 95.2   \\ \hline
\multicolumn{1}{|c|}{\multirow{3}{*}{ImageNet}} & ResNet-18             & \multicolumn{1}{c|}{71.4}  & \multicolumn{1}{c|}{67.8} & \multicolumn{1}{c|}{70.9} & 71.2   \\ \cline{2-6} 
\multicolumn{1}{|c|}{}                          & RegNetX-400MF          & \multicolumn{1}{c|}{73.3}  & \multicolumn{1}{c|}{64.3} & \multicolumn{1}{c|}{72.1} & 72.4   \\ \cline{2-6} 
\multicolumn{1}{|c|}{}                          & ResNet-50              & \multicolumn{1}{c|}{78.1}  & \multicolumn{1}{c|}{70.3} & \multicolumn{1}{c|}{77.7} & 77.6  \\ \hline
\end{tabular}
\end{center}
\label{tab:accuracy}
\end{table}

As discussed earlier in Section~\ref{sec:sdgp:motiv}, rescaling the remaining gradients makes it so that pruning does not pull the average towards zero. We speculate that this may also have some interaction with Batch Normalization~\cite{ioffe2015batch}, as it uses mean and variance statistics across each batch. The rescaling operation ensures that the variance of data gradients is preserved across each sample and feature map. 

\subsection{Impact of Sparsity on Accuracy}
\label{sec:eval:sparsity}
While the A100 only supports a ($n=2$ nonzero, $m=4$ group size) structured sparsity pattern, we are interested in seeing how varying $n$ and $m$ impact final model accuracy. Since the 2:4 sparsity setting leads to a $2\times$ speed-up when using the sparse tensor cores compared to using the dense tensor cores, we assume that even sparser settings would lead to additional reductions in computation time. 

\begin{figure}
    \includegraphics[width=0.9\columnwidth]{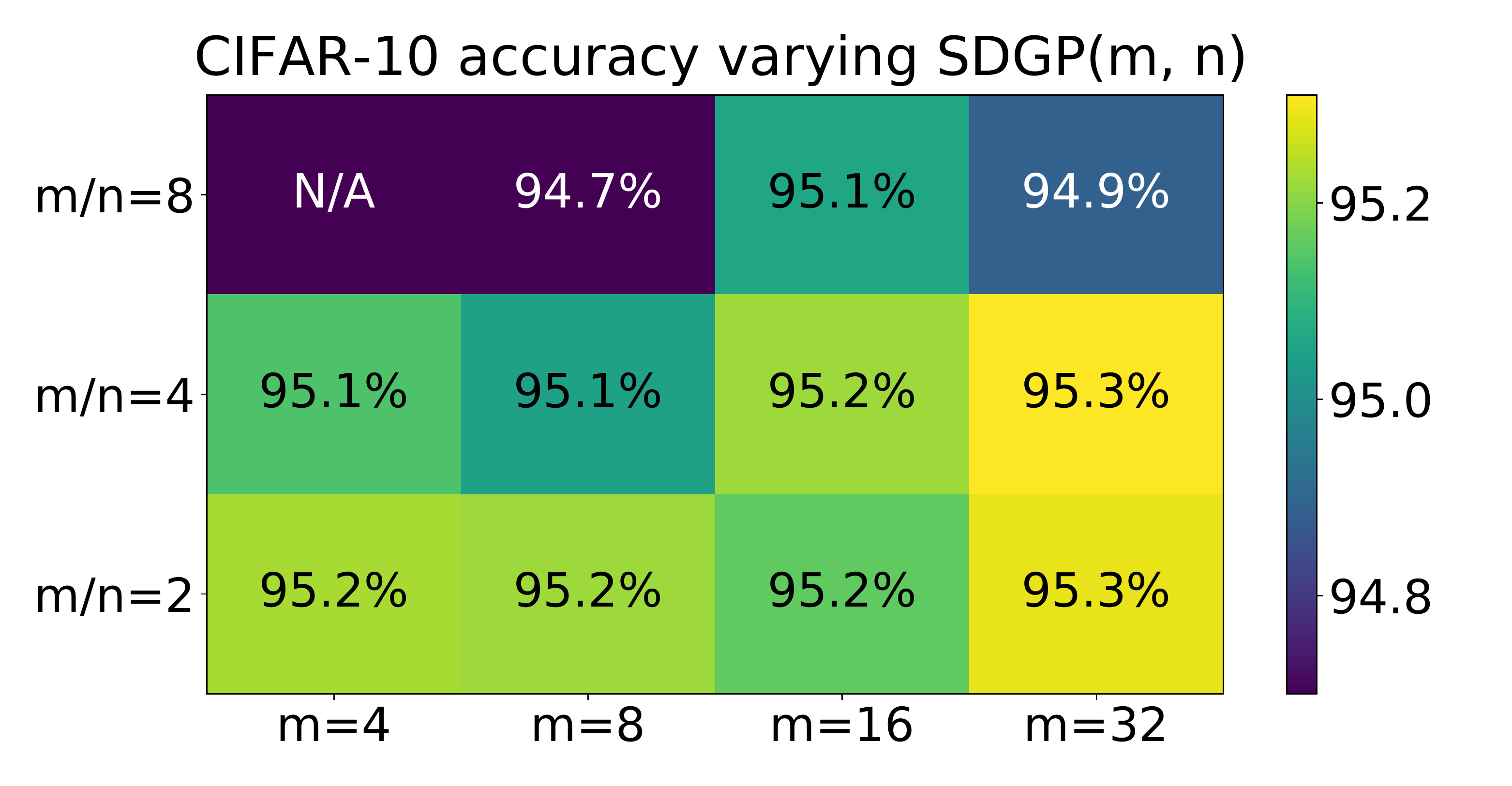}
    \caption{The final top-1 validation accuracy for ResNet-0 on CIFAR-10 for different $n$ and $m$ configurations. Each row represents a sparsity ratio ($r = \nicefrac{m}{n}$). Each column corresponds to a group size $m$. Note that the upper-left cell ($\nicefrac{m}{n} = 8, m=4$) is labeled N/A as it is not possible.}
    \label{fig:sparse-settings}
\end{figure}

Figure~\ref{fig:sparse-settings} shows how the classification accuracy of CIFAR-10 using ResNet-9 changes as a function of $n$ and $m$. All settings are evaluated using the Rescaled Magnitude pruning function. As the sparsity ratio ($r = \nicefrac{m}{n}$) increases, the classification accuracy decreases as expected. However, there are two interesting insights we can take from this data. 

First, for a fixed sparsity ratio (e.g., $2$), we see that using a larger group size $m$ leads to a higher classification accuracy. This is due to the added flexibility in which nonzero elements are selected when using a larger group size. For example, picking the largest $n=2$ nonzeros for $m=4$ is less flexible than picking the largest $n=4$ nonzeros for $m=8$. This added flexibility leads to less error introduced by pruning. Second, Figure~\ref{fig:sparse-settings} suggests that a higher sparsity ratio, such a $4$, could be used assuming that a larger group size $m$ could be support in hardware. In the next section, we discuss how this increased sparsity ratio would translate into a training time speedup.

\subsection{Reduction in DNN Training Time}
\label{sec:eval:reduction}
In this section, we analyze the potential reduction in training time when using SDGP. Note that, while sparse tensor cores in the A100 can perform matrix multiplication $2\times$ faster than dense tensor cores, their use is not currently supported in deep learning frameworks such as PyTorch. Therefore, we estimate the speedup gained via SDGP by measuring the running time of the forward and backward passes for each convolution layer in a given network and reducing the computation time of computing data gradients based on the sparsity ratio introduced by SDGP.

\begin{figure}
    \includegraphics[width=\columnwidth]{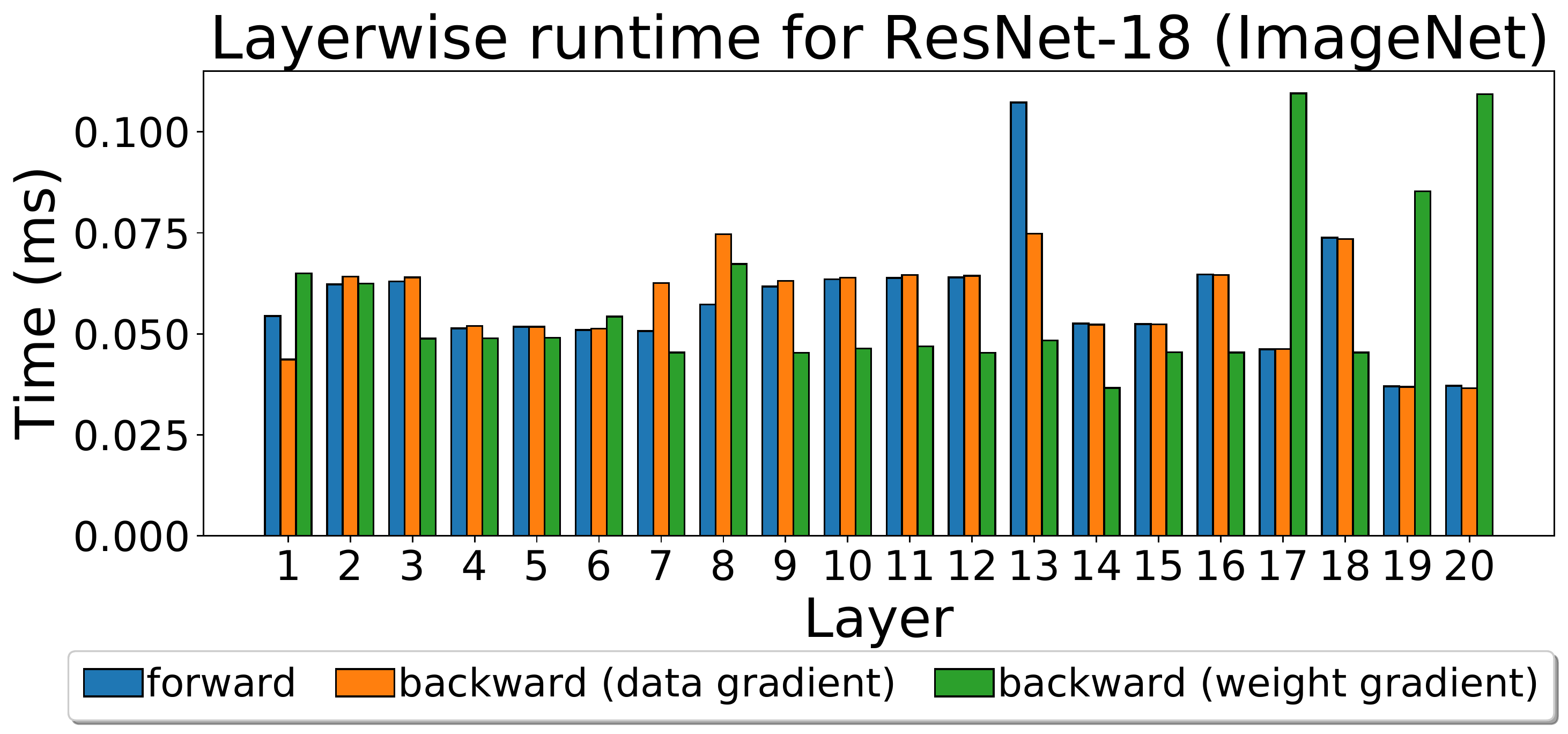}
    \caption{The running time in milliseconds for the forward and backward computations for each convolution layer in ResNet-18 for a batch of samples with dimensions $B=256, C=3, W=224, H=224$. SDGP reduces the backward (data gradient) computations, which account for approximately one-third of the runtime.}
    \label{fig:layer-runtimes}
\end{figure}

Figure~\ref{fig:layer-runtimes} shows the running time for all convolutions per layer (1 in forward pass and 2 in backward pass corresponding to Figure~\ref{fig:dnn-training}). Each grouping shows the total running time for a convolution layer (forward pass, backward data gradients, backward weight gradients). Adding all the times up, we see that computing data gradients takes approximately one-third of the total running time (33.17\% in this case for ResNet-18). Thus, applying SDGP with $n=2$ and $m=4$ (sparsity ratio of 2) should lead to a 16.6\% reduction in total runtime. Similarly, a sparsity ratio of 4 would lead to a 24.9\% runtime reduction.

\begin{figure}
    \includegraphics[width=\columnwidth]{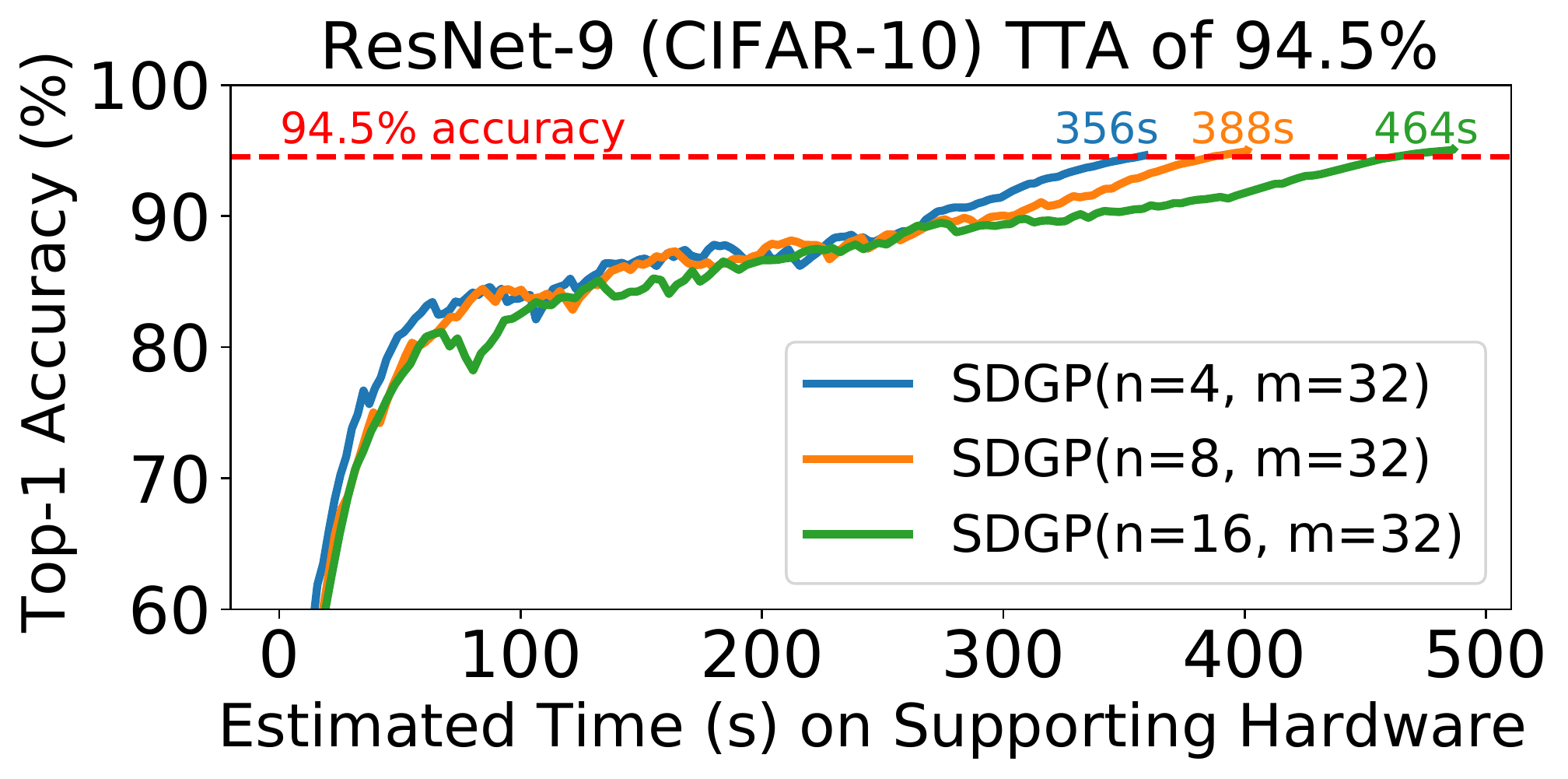}
    \caption{The estimated Time-To-Accuracy (TTA) of 94.5\% for the ResNet-9 model trained with different SDGP settings. The running time was estimated by taking the real running time and subtracting the savings due to SDGP when performing the data gradient updates.}
    \label{fig:tta}
\end{figure}

Figure~\ref{fig:tta} shows the estimated reducing in training time to meet a target accuracy 94.5\% for different SDGP sparsity settings on CIFAR-10. Since the Nvidia A100 does not support N:M sparsity settings other than $n=2$, $m=4$, we estimated these run times by reducing the fraction of time spend performing the data gradient computations based on the SDGP sparsity ratio. For instance, for SDGP($n=4$, $m=32$), which has a sparsity ratio of $\nicefrac{32}{4} = 8$, we compute the reduced data gradient time by taking the original percentage of data gradient computation time ($33.31\%$) and dividing it by the sparsity ratio to get the new computation percentage for data gradients $\nicefrac{33.31\%}{8} = 4.16\%$. We then rescale the original training time recorded per epoch to account for this reduction.

After estimating the runtime for each SDGP setting, we observe that the highest sparsity setting of SDGP($n=4$, $m=32$) outperforms the other settings. Specifically, it is able to achieve a relative speedup of 27\% compared to the less aggressive sparsity setting of SDGP($n=16$, $m=32$) for the same TTA. This suggests that it may be useful to have more fine-grained hardware support for varying levels of structured sparsity (e.g., 1:8, 2:8, 4:8), in order to enable an efficient trade-off between training time and accuracy.

\section{Future Work and Conclusion}
\label{sec:conc}
In this paper, we proposed a novel structured data gradient pruning (SDGP) technique to speed up DNN training. The structured pruning performed by SDGP is naturally supported by modern hardware accelerators like the Nvidia A100 GPU. SDGP enables a 15-25\% reduction in DNN training time without a significant impact to classification performance. We evaluated SDGP on multiple CNNs (ResNet-18, ResNet-50, and RegNetX-400MF) on ImageNet and demonstrate that it can achieve comparable performance to the dense baseline (within 0.3\% across all models).

We also investigated the impact of higher structured sparsity ratios on the reduction in training time and impact to performance. For some tasks, we believe that adding support for multiple structured sparsity settings will faster training time while maintaining acceptable accuracy. For instance, training a CIFAR-10 model using SDGP($n=4$, $m=32$) leads an additional 27\% speed up over SDGP($n=2$, $m=4$) while for a 0.4\% reduction in accuracy.

Once structured sparsity is better supported by modern deep learning frameworks, we hope that SDGP can be easily to add to existing training pipelines in a similar fashion to FP16 support. As illustrated, in some cases, this would lead to a relatively significant reduction in training time without much effort for the user (no modifications to learning rate, model structure, or other hyperparameters).

\bibliographystyle{plain}
\bibliography{references}

\end{document}